\documentclass{article}
\usepackage{spconf,amsmath,graphicx}

\usepackage{enumitem}
\usepackage{amsmath,graphicx}
\usepackage[utf8]{inputenc}
\usepackage[english]{babel}
\usepackage{amssymb}

\usepackage{color}
\usepackage{url}
\usepackage{caption}
\usepackage{multirow}
\usepackage{textcomp}
\usepackage{booktabs}
\usepackage{multicol}
\usepackage{algorithm2e}
\usepackage{colortbl} 
\usepackage{todonotes}
 \usepackage[nolist,nohyperlinks]{acronym}

\def\x{{\mathbf x}}
\def\L{{\cal L}}

\usepackage[acronym]{glossaries}
\acrodef{mIoU}{mean intersection over union}
\acrodef{DSC}{dice coefficient}
\acrodef{CT}{computed tomography}

\setlist{nosep, leftmargin=14pt}

\usepackage{mwe} 

\def\x{{\mathbf x}}
\def\L{{\cal L}}

\title{Is Long Range Sequential Modeling Necessary for Colorectal Tumor Segmentation?}
\name{Abhishek Srivastava$^1$, Koushik Biswas$^1$, Gorkem Durak$^1$, Gulsah Ozden$^2$, Mustafa Adli$^2$, Ulas Bagci$^1$}
\address{$^1$Machine and Hybrid Intelligence Lab, Northwestern University, Chicago, IL, USA.\\$^2$Department of Radiation Oncology, Marmara University School of Medicine, Turkey.}
\begin{document}
%

\maketitle
\def\x{{\mathbf x}}
\def\L{{\cal L}}
\begin{abstract}
Segmentation of colorectal cancer (CRC) tumors in 3D medical imaging is both complex and clinically critical, providing vital support for effective radiation therapy planning and survival outcome assessment. Recently, 3D volumetric segmentation architectures incorporating long-range sequence modeling mechanisms, such as Transformers and Mamba, have gained attention for their capacity to achieve high accuracy in 3D medical image segmentation. In this work, we evaluate the effectiveness of these global token modeling techniques by pitting them against our proposed MambaOutUNet within the context of our newly introduced colorectal tumor segmentation dataset (CTS-204). Our findings suggest that robust local token interactions can outperform long-range modeling techniques in cases where the region of interest is small and anatomically complex, proposing a potential shift in 3D tumor segmentation research.
\end{abstract}
\section{Introduction}
\label{sec:intro}

Colorectal cancer (CRC) tumor segmentation from medical scans, such as computed tomography (CT), is a crucial task in the field of medical imaging and computer-assisted diagnosis. However, this task presents several challenges, variability in image quality and the complex anatomy of the colorectal region hinders precise delineation of tumors. Many CRC tumors are small, irregular, or have complex shapes, making accurate segmentation diffcult. The presence of multiple lesions or tumors in the same patient further adds complexity to the tasks.

To overcome these challenges, researchers and clinicians employ various deep learning techniques to segment CRC tumors. For instance, convolutional neural networks (CNNs)~\cite{ronneberger2015u} and, more recently, Transformers~\cite{cao2022swin,hatamizadeh2021swin}, have proven to be effective for 3D image segmentation, with CNNs capturing local features efficiently and Transformers modeling long-range dependencies~\cite{dosovitskiy2020image}. However, the self-attention mechanism~\cite{vaswani2017attention} of Transformers suffers from quadratic complexity bottleneck, resulting in significant computational demands. 
This challenge was addressed by the Mamba~\cite{gu2023mamba} architecture inspired by state space models, which facilitates long-range dependency modeling via a selection mechanism and a hardware-efficient algorithm. U-Mamba~\cite{ma2024u} integrates Mamba into nnU-Net~\cite{isensee2021nnu}. SegMamba~\cite{xing2024segmamba} proposed a tri-oriented Mamba (ToM) module for enhanced 3D feature modeling. MambaOut~\cite{yu2024mambaout} analyzes Mamba's application computer vision tasks, hypothesizing that while it has advantages in segmentation/object detection due to their long range modeling requirements, Mamba has limitations due to their inability to model non-causal features. In this paper, we provide a comprehensive comparison of these different local- and global token modeling mechanisms on our newly proposed Colorectal Tumor Segmentation 204 (CTS-204) dataset and instantiate a new methodology MambaOutUnet to provide strong evidence for the hypothesis introduced in~\cite{yu2024mambaout}
Our contributions are as follows:\\ 
\textbf{1) Is Long-Range Sequence Modeling Necessary for Tumor Segmentation?} While image segmentation is essentially a long range sequence modeling task, in case of tumor segmentation where the target region by its natural properties has no correlation with most of the surrounding pixels/voxels, does it still holds the same efficacy? We investigate the potential benefits and limitations of this approach in 3D tumor segmentation.\\
\textbf{2) We propose a newly curated CTS-204 dataset for colorectal cancer tumor segmentation}.\\
\textbf{3) Comprehensive Analysis of Long Range Sequence Modeling Architectures:} We propose a comprehensive analysis of recent Mamba-based architectures, including UNet variations that enhance the capacity to model local- and global token interactions. Our results demonstrate that efficient channel mixing and spatially gated features can outperform many existing computationally intensive long-range modeling techniques. Specifically, we analyze the performance of MambaOutUNet on our CTS-204 dataset and compare it to established 3D segmentation architectures.
\begin{table*}[!t]
\centering
\footnotesize
\setlength{\tabcolsep}{18pt}
\renewcommand{\arraystretch}{1.1}
\caption{Quantitative Comparison on the CTS 204 and Synapse multi-organ CT dataset (BTCV). We report the Dice Score (DSC), Mean Intersection over Union (mIoU), and Normalized Surface Distance (NSD)}
\begin{tabular}{@{}l|ccc|ccc@{}}
\toprule
\multirow{2}{*}{\textbf{Method}} & \multicolumn{3}{c|}{\textbf{CTS 204}} & \multicolumn{3}{c}{\textbf{BTCV}} \\
 & \textbf{DSC} & \textbf{mIoU} & \textbf{NSD} & \textbf{DSC} & \textbf{mIoU} & \textbf{NSD} \\ 
\hline
UNet & \textcolor{magenta}{0.5007} & \textcolor{magenta}{0.3839} & \textcolor{magenta}{0.6116} & \textcolor{magenta}{0.8217} & \textcolor{magenta}{0.6982} & \textcolor{magenta}{0.9320} \\ \hline
nnU-Net & 0.4842 & 0.3619 & 0.6073 & 0.7775 & 0.6351 & 0.9244 \\ \hline
SwinUnet & 0.4602 & 0.3388 & 0.5060 & 0.8023 & 0.6718 & 0.8982 \\ \hline
VisMix & 0.3612 & 0.2566 & 0.4663 & 0.7436 & 0.5906 & 0.8025 \\ \hline
SegMamba & 0.3990 & 0.2982 & 0.5256 & 0.8184 & 0.7124 & 0.9158 \\ \hline
SegHydra & 0.3687 & 0.2728 & 0.5972 & 0.7917 & 0.6803 & 0.8823 \\ \hline
MambaOutUNet & \textbf{0.5203} & \textbf{0.3950} & \textbf{0.6139} & \textbf{0.8338} & \textbf{0.7345} & \textbf{0.9397} \\ 
\bottomrule
\end{tabular}
\label{tab:result_combined}
\end{table*}

\begin{figure*}[!t]
    \centering
    \includegraphics[width=0.9\textwidth]{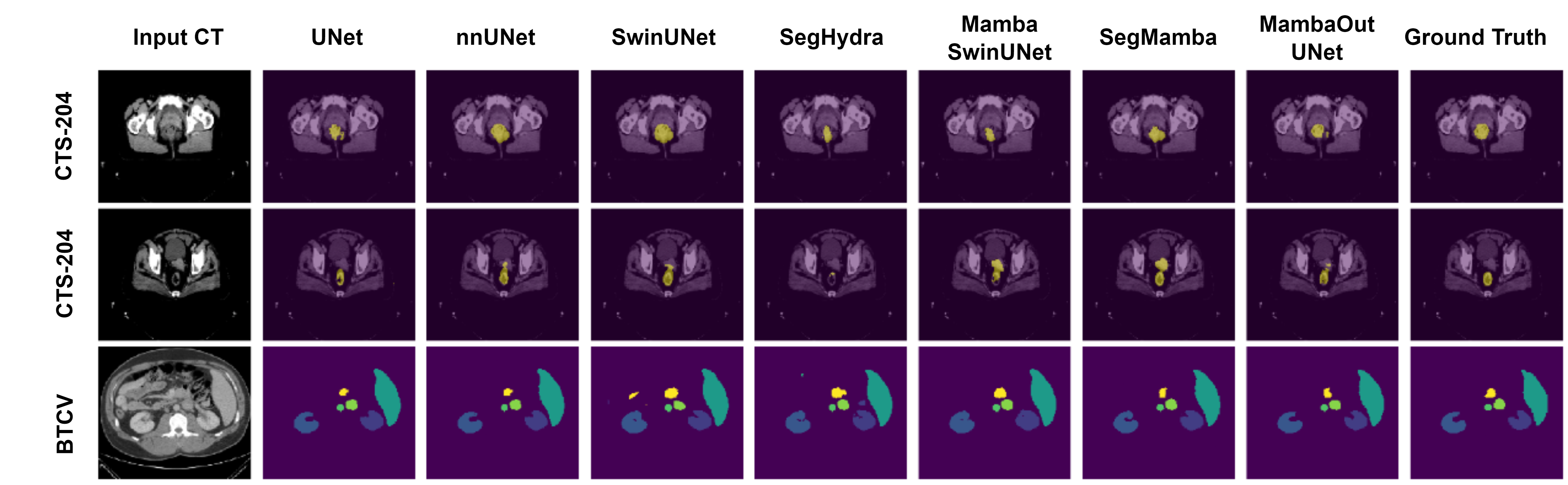}
    \caption{Qualitative comparison of MambaOutUNet with other established and proposed baselines.}
    \label{fig:qualitative}
    \vspace{-5mm}
\end{figure*}
\section{Dataset}
We have publicly released a new colorectal tumor segmentation dataset termed "CTS-204". The dataset was collected from Marmara University Pendik Research and Education Hospital in Turkey and includes CT scans from 204 patients undergoing treatment for colon cancer. All sensitive patient information has been de-identified. Two experienced radiation oncologists annotated each volume in consensus. All the CT scans were acquired with tube voltage, 120 Kv, pitch, 0.85-1.2, Z-axis spacing from 3.0 mm to 3.75 mm, with a median of 3.75 mm. CTS-204 will be made available upon request. Additionally, to validate the performance of our MambaOut method and our hypothesis that long-range sequence modeling is still useful for segmentation problems where global token modeling is informative, we include \textit{Synapse multi-organ segmentation dataset}.
\vspace{-3mm}
\section{Methods}
\vspace{-3mm}
In this section, we describe the architecture of our three proposed baselines: Mamba-SwinUnet, SegHydra, and MambaOutUnet. Assessing the impact of long range sequence modeling through Mamba requires a fair comparison with SegMamba~\cite{xing2024segmamba}. We ensure this by following the same architecture as proposed in SegMamba~\cite{xing2024segmamba} while invoking necessary architectural changes in the \textbf{encoder} for each proposed technique. The architecture of Mamba-SwinUnet, SegHydra, and MambaOutUnet along with the Tri-oriented spatial Mamba (TSMamba) block and the convolutional layer incorporated 3D decoder will be introduced in the following section. 
\subsection{Encoder}
The 3D input volume \( CT \in R^{C \times D \times H \times W} \) is initially passed through the encoder stem comprising a depth-wise convolutional layer with a kernel size of \(7 \times 7 \times 7\), padding of \(3 \times 3 \times 3\), and stride of \(2 \times 2 \times 2\). The resultant \( x_0 \in \mathbb{R}^{48 \times \frac{D}{2} \times \frac{H}{2} \times \frac{W}{2}} \) tensor is sent to the subsequent TSMamba Blocks and downsampling layers. \textbf{Tri-orientated Spatial Mamba (TSMamba) Block:}
TSMamba blocks facilitate global token modeling through Mamba and their hierarchal nature in the UNet allows modeling of multi-scale features. Global token modeling and capturing multi-scale features are conducive to obtaining precise segmentation maps. Specifically, the input feature in $lth$ layer of TSMamba block are first passed through the Gated Spatial Convolutional module. This module consists of two convolutional blocks each consisting of normalization, convolution and non-linear activation layer. The two convolutional layers have kernel size of \(3 \times 3 \times 3\) and \(1 \times 1 \times 1\), respectively. The resultant outputs undergo an element wise multiplication so as to regulate the amount of informative spatial features and suppress irrelevant features, similar to a gated attention mechanism(see Equation~\ref{eq:gsc_def}, where C is the convolutional layers and x are the input feature maps)
\begin{equation}
    \text{GSC}(x) = x + C_{3 \times 3 \times 3} \left( C_{3 \times 3 \times 3}(x) \cdot C_{1 \times 1 \times 1}(x) \right), \label{eq:gsc_def}
\end{equation}
Since Mamba layers performs token-to-token modeling in one direction the input 3D features are flattened into three different sequences corresponding to the axial, coronal, and sagittal plane(see Equation~\ref{eq:tri_orientated_mamba}).
\begin{equation}
\text{ToM}(x) = \text{Mamba}(x_{\text{axial}}) + \text{Mamba}(x_{\text{coronal}}) + \text{Mamba}(x_{\text{sagittal}})
\label{eq:tri_orientated_mamba}
\end{equation}
Subsequently, a multi-layer perceptron (MLP) is used to perform channel mixing and layer normalization is used to further model feature representations. The information flow can be summarized as:
\begin{align}
    \hat{x}^l &= \text{GSC}(x^l), \label{eq:gsc} \\
    \tilde{x}^l &= \text{ToM} \left( \text{LN} \left( \hat{x}^l \right) \right) + \hat{x}^l, \label{eq:tom} \\
    x^{l+1} &= \text{MLP} \left( \text{LN} \left( \tilde{x}^l \right) \right) + \tilde{x}^l, \label{eq:mlp}
\end{align}
\subsection{Decoder and Feature-level Uncertainty Estimation (FUE)}
The encoder generates multi-scale features which include embedded uncertainty information~\cite{zhao2023uncertainty}. These feature maps are passed through FUE~\cite{xing2024segmamba} that stabilizes features with lower uncertainty levels within each skip connection, enhancing the reliability of multi-scale feature representations. The decoder progressively refines feature maps from the bottleneck layer using a series of upscaling blocks. Each decoder block doubles the spatial resolution and combines multi-scale features from the encoder at each level \( x^i \). The transformation in each block is defined as:
\begin{equation}
D_x = \text{ReLU} \left( \text{Norm} \left( \text{Conv}_{3 \times 3} \left( \text{ConvT}(D_{x-1}) \oplus \text{FUE}(x^i) \right) \right) \right)
\label{eq:decoder_block}
\end{equation}
where \( D_0 \) is the output from the first decoder block. Here, \(\text{ConvT}\) represents a 3x3 transposed convolution with stride 2, \(\text{Norm}\) denotes instance normalization, and \(\oplus\) is the concatenation operation. The final segmentation head uses a convolutional layer on \( D_1 \) to produce the output with the required number of class channels.
\vspace{-4mm}
\subsection{SegHydra}


Hydra is a sequence model that incorporates a \textbf{quasiseparable matrix mixer} for effective bidirectional sequence processing. Quasiseparable matrices offer an ideal structure for sequence modeling by generalizing low-rank mixers in linear attention and semiseparable matrices in state-space models (SSMs). This structure provides enhanced computational efficiency while preserving high model expressivity.

The quasiseparable matrix structure allows bidirectional sequence mixing holding an advantage over SSMs for its application in non-causal scenarios. SSMs rely on semiseperable matrices for sequence modeling~\cite{dao2024transformers} and can overcome this limitation by using two SSMs for forward and backward sequence modeling, combining thier features through addition/concatenation. In contrast, Hydra uses quasiseperable matrices to allow non-causal modeling posting three advantages a.) better representation power, b.) sub-quadratic matrix multiplication, c.) lower parameters as compared to alternatives.
A matrix \( M \) is defined as \( N \)-quasiseparable if its elements \( m_{ij} \) satisfy the following conditions:

\begin{equation}
    m_{ij} = 
    \begin{cases} 
      \overrightarrow{c_i}^T \, \overrightarrow{A_{i:j}} \, \overrightarrow{b_j}, & \text{if } i > j \\
      \delta_i, & \text{if } i = j \\
      \overleftarrow{c_j}^T \, \overleftarrow{A_{j:i}} \, \overleftarrow{b_i}, & \text{if } i < j
   \end{cases} \label{eq:quasi_matrix_definition}
\end{equation}

Here, \( \delta_i \) is a scalar, and \( \overrightarrow{b_j}, \overrightarrow{c_i}, \overleftarrow{b_i}, \overleftarrow{c_j} \in \mathbb{R}^{N \times 1} \), with \( A_i \in \mathbb{R}^{N \times N} \). This formulation permits bidirectional processing by including non-zero entries in the upper triangular section.
Tri-oriented Mamba block in SegMamba, processes flattened sequence across three different views of the same 3D input volume through mamba blocks. We reduce parametric complexity and could increase the efficiency of long range modeling by using the quasiseparable matrix multiplication to provide a more effective and efficient alternative while maintaining other aspects of the encoder-decoder architecture described earlier. This operation in Hydra can be efficiently computed by decomposing it into operations involving semiseparable matrices. 
\begin{equation}
    \text{QS}(x_l) = \text{shift}(\text{SS}(x_l)) + \text{flip}(\text{shift}(\text{SS}(\text{flip}(x_l)))) + Dx_l, \label{eq:quasi_multiplication}
\end{equation}

where:
\(\text{QS}(x_l) \): quasiseparable operation on input \( x_l \), \( \text{SS}(x_l) \): semiseparable operation on \( x_l \), \( \text{flip}(x_l) \): reverses \( x_l \), \( \text{shift}(x_l) \): shifts \( x_l \) right by one position with zero-padding at the start, and \( D = \text{diag}(\delta_1, \ldots, \delta_L) \): diagonal matrix with parameters \( \delta_i \). 
This decomposition enables Hydra to leverage efficient linear-time semiseparable matrix multiplications, compatible with various SSMs, for high-performance sequence modeling.
\vspace{-4mm}
\subsection{Vision Mamba Block with Swin Transformer Integration}
MambaVision~\cite{hatamizadeh2024mambavision} empirically showed the advantage of combining Mamba layers succeeded by the self-attention mechanism. Mamba’s auto regressive structure, while effective for sequence modeling, presents limitations in tasks like segmentation and object detection that benefit from a global receptive field. Specifically:
a.) Image pixels do not exhibit strict sequential dependencies, making Mamba’s step-wise approach less efficient for spatial data.
b.) Autoregressive processing limits the model’s ability to capture global context in a single pass, essential for many vision tasks.
However, using self-attention following Mamba layers for 3D volumes can be computationally heavy and not feasible to train. To address these challenges, we propose a restructured Mamba-SwinUNet that integrates Swin-Transformer blocks~\cite{hatamizadeh2021swin} for improved spatial context representation. In our design, we augment each Mamba block with a Swin Transformer layer after every Mamba processing layer. This hybrid structure balances Mamba’s sequential strengths with the Swin Transformer's ability to capture global and local spatial dependencies more effectively. For the \( l \)-th Vision Mamba block, the computation is given by \( x_{l} = \text{Swin}(\text{LN}(\text{Mamba}(x_l))) + x_l \), where \( \text{Mamba}(x_l) \) processes the input \( x_l \) to capture sequential dependencies, \( \text{Swin}(\cdot) \) applies the Swin Transformer block to model spatial context and improve receptive field coverage, and \( x_l \) denotes the original input feature at layer \( l \). This hybrid Mamba-SwinUNet approach combines the Mamba model's sequential feature extraction capabilities with the Swin Transformer's global spatial learning, thus enhancing both local and global feature understanding.
\vspace{-2mm}
\subsection{MambaOutUNet}
To empirically validate our hypothesis, we follow the protocol set in MambaOut~\cite{yu2024mambaout}, the Mamba block is analyzed alongside the Gated CNN block~\cite{dauphin2017language}. 
Given input \( X \in \mathbb{R}^{N \times D} \), the Mamba block meta-architecture integrates token mixing, represented by a token mixer, with normalization and an MLP. In this setup:
- \( \text{Norm}(\cdot) \) denotes the normalization layer.
- \( \text{TokenMixer}(\cdot) \) performs token mixing, enhancing spatial feature extraction.
For token mixing, the Gated CNN and Mamba differ as follows:
\begin{equation}
    \text{TokenMixer}_{\text{GatedCNN}}(x) = \text{Conv}(x), \label{eq:token_mixer_gatedcnn}
\end{equation}
\begin{equation}
    \text{TokenMixer}_{\text{Mamba}}(x) = \text{SSM}(\sigma(\text{Conv}(x))), \label{eq:token_mixer_mamba}
\end{equation}
where \( \sigma \) is the activation function and SSM represents the sequential state-space model for improved token mixing.
To isolate the impact of SSM, we use a simplified methodology, MambaOut, based on the Gated CNN block without SSM. This setup is incorporated inside the MambaOutUNet, specifically inside the encoder where we ablate the Mamba layers and use GatedCNN instead. Thus, allowing us to accurately assess the impact of Mamba on segmentation performance through our experiments.
\section{Experiments}
CTS-204 was split into 3 splits, training, validation and testing which had 163, 20, 21 cases, respectively. The relevant baselines and all proposed architectures were trained on CTS-204 for 200 epochs for the same learning rate and validation interval. Similarly, all methodologies were trained on the BTCV dataset for 400 epochs on a single Tesla A100 GPU. Following~\cite{pitfallmetric} we use three metrics DSC, mIoU, and NSD to evaluate performance.
\section{Results and Discussion}
To validate our hypotheses, we conducted experiments on two datasets: the CTS 204 dataset and the BTCV multi-organ segmentation dataset. The results of these experiments are summarized in Tables \ref{tab:result_combined}. Here we can observe that the MambaOutUNet achieves a (DSC), mIoU, and NSD of 0.5203, 0.6971, and 0.6971 outperforming the second best benchmarks by 1.96\%,1.11\%,0.23\%, respectively. Moreover, a performance gain of 15.19\% and 9.68\% is observed over the SegMamba~\cite{xing2024segmamba}, this can be attributed to the fact that even though segmentation is essentially a \textit{long-sequence task}, tumor segmentation requires focus on immediate locality structure surrounding cancerous tissues for accurate delineation, thus, Mamba, Transformer  and other long-range sequence modeling techniques have limited use in such cases. It should also be noted that while Hydra's quasisemiseperable matrix mixing technique holds various advantages over bidirectional mamba, it is outperformed by SegMamba on both CTS-204 and BTCV, suggesting that, despite their claims~\cite{hwang2024hydra}, inherent RNN nature of SSMs limits modeling of non-causal features. Methods incorporating SSMs and Transformer  architectures, while beneficial in contexts requiring long-range dependencies, do not provide the same level of performance for localized segmentation tasks. The focus on global context in these models may lead to unnecessary complexity without yielding proportional improvements in accuracy.

However, From Table~\ref{tab:result_combined} and Figure~\ref{fig:qualitative}, we can observe that on BTCV multi-organ segmentation challenge, where the region-of-interest has high variance in size and structure, long range sequencing is desirable, thus, Mamba and Transformer  based models show competitive performance. Therefore, even though long range sequencing might be desirable in medical image segmentation and is worthwhile exploring, there must also be a need to study these new architectures equipped with Mamba and self-attention variants in the context of datasets with small regions of interest and complex structural variations, such as the newly introduced CTS-204. To this extent, MambaOutUNet can serve as a useful baseline to asses the performance gain obtained through long-range sequencing. We would also like to highlight that there should be an increased effort to design architectures specifically for more efficient local token modeling to capture the small, complex and difficult to observe ROIs like colorectal tumors in medical image analysis.
\vspace{-4mm}
\section{CONCLUSION}
\vspace{-2mm}
In this paper, we release CTS-204, a new colorectal tumor segmentation dataset with 204 distinct cases. We demonstrate the effectiveness of our MambaOut architecture for colorectal tumor segmentation which provides critical insights and a robust framework for future research in tumor segmentation. Our findings suggest a promising direction for further investigation into segmentation methodologies that prioritize performance on small-scale features. Additionally, we critically analyze segmentation architectures embedded with recent long range sequencing techniques and reassess their performance on two different medical image segmentation, which empirically supports our hypothesis. Moreover, we establish the importance of MambaOutUnet as a baseline architecture for future 3D volumetric segmentation research. 


\section{Compliance with ethical standards}
\label{sec:ethics}
This study was performed in line with the principles of the Declaration of Helsinki. Approval was granted by Northwestern University (No. STU00214545).

\section{Acknowledgments}
\label{sec:acknowledgments}
This project is supported by NIH funding: R01-CA246704, R01-CA240639, U01-DK127384-02S1, and U01-CA268808.

\bibliographystyle{IEEEbib}
\bibliography{strings,refs}

\end{document}